# Aspect-Based Sentiment Evolution and its Correlation with Review Rounds in Multi-Round Peer Reviews: A Deep Learning Approach

Ruxue Han[a,b], Haomin Zhou[a,b], Jiangtao Zhong[a], Chengzhi Zhang[a,*]

[a] *Department of Information Management, Nanjing University of Science and Technology, Nanjing 210094, People's Republic of China*

[b] *School of Economics and Management, East China Normal University, Shanghai 200062, People's Republic of China*

**Abstract:** Mining sentiment information from the textual content of peer review comments offers valuable insights into the scientific evaluation process. However, previous studies are often constrained by coarse-grained analysis and the lack of differentiation across review rounds. Notably, the dynamic shifts in reviewers' focus and sentiment tendencies throughout multiple review stages remain underexplored. To address this gap, the present study investigates the distribution and evolution of aspect-level sentiments and examines their correlation with the number of review rounds. We begin by segmenting the multi-round review comments of 11,063 accepted papers from Nature Communications and identifying fine-grained review aspect clusters. A manually annotated corpus of approximately 5,000 review sentences is then constructed. Using this dataset, we train a series of deep learning-based aspect sentiment classification models. Among them, the LCF-BERT-CDM model achieves the best performance, with a Macro-$F_1$ score of 82.65%. Subsequent statistical analysis reveals a consistent trend: as the number of review rounds increases, the proportion of positive sentiments rises, while negative sentiments decline. Correlation analysis further indicates that aspect sentiment scores are negatively associated with the total number of review rounds. Key aspects exhibiting stronger correlations include "experiments", "research significance" and "result analysis".

---

[*] Corresponding author: Chengzhi Zhang
E-mail address: zhangcz@njust.edu.cn

## 1 Introduction

Peer review is a process in which experts review the quality of research work in their field. Currently, the peer review mechanism has been adopted by most journals and conferences, and plays an important role in the scientific writing and publishing process (Kang et al., 2018). Open Peer Review (OPR), an important component of open science, emphasizes the transparency of manuscript review reports as a key element. This enhances research transparency and offers data for text mining analysis of peer review comments.

Peer review comments on academic articles not only include reviewers' overall evaluations but also specific opinions on strengths and weaknesses of the article, providing a rich source of sentiment information (Chakraborty et al., 2020). A review comment for a manuscript is organized around several key aspects, which typically encompass critical sections, content, and the aspects of the article mentioned by reviewers within their comments regarding the article. Furthermore, as journal articles are typically polished through multiple review rounds from submission to publication, the focus and sentiment tendencies expressed by reviewers might change across these rounds. To further understand the reviewers' views of articles, the fine-grained sentiment information hidden in multiple rounds of review comments can be mined through sentiment analysis.

At present, scholars have conducted content mining research of peer review comments focusing on sentiment analysis. However, a limitation exists in the coarse granularity of aspects like innovation and clarity, offering an overall evaluation without capturing reviewers' specific sentiments towards the evaluated object. This lacks precise guidance for authors in optimizing their articles. In addition, most studies on peer review comments do not distinguish the rounds of review, and rarely pay attention to the characteristics of the review rounds, because general journal papers typically only provide information about the overall review cycle reflecting the time duration, but rarely specifying the number of review rounds. Focusing on the rounds can divide the whole review process into each round. And the authors constantly revise and polish the articles based on the feedback of review comments, therefore, the key aspects that the reviewers pay attention to and the associated sentiment tendencies of the aspects across multiple review rounds may change. Figure 1 (a) ~ (b) illustrates a sample of three review rounds for an article. In the first round of review, reviewers evaluated the aspects of the findings, charts and result analysis, expressing more negative sentiment in their comments on analysis and reasoning interpretation. In the third round (the last round) review, experts only highlighted an icon problem, negative sentiment decreased with the increase of the review round. Using sentiment analysis to mine review comments across various rounds can reveal the sentimental tendency and changes of the aspects of the article under different review rounds, which helps us to further understand the peer review process of academic articles from submission to publication.

Review Aspect　Opinion Word

Reviewers' Comments:

Reviewer #1:
Remarks to the Author:
This is an interesting report characterizing a new mouse strain with a novel phenotype. Parts of the report present convincing new findings regarding an important role for TRAF3 in regulation of osteoblast differentiation. The final Figures are also straightforward and reasonable. However, some portions (especially Figs. 3 and 4) are problematic, over-interpreted, and in some instances, based upon flawed reasoning. Specific comments are as follows.

（a）Excerpts of sample comments from the first round of review

Reviewers' Comments:

Reviewer #3:
Remarks to the Author:
The authors have adequately addressed my criticisms.
One minor edit: add WT and TRAF3KO labels to figure 4L.

（b）Sample comments from the third round of review

**Figure 1 Multiple rounds of peer review comments samples** [1]

Therefore, this paper intends to conduct sentiment analysis on peer review comments in multiple rounds at a finer-grained aspect level. Taking the peer review comments corpus of articles accepted in the Nature Communications journal for example, this paper will address the following three research questions:

***RQ1***: What is the sentiment distribution of aspects in multiple rounds of peer review comments?

This involves investigating the distribution of positive, neutral, and negative sentiment polarities within review comments, identifying aspects with higher sentiment distribution to uncover reviewer focus.

***RQ2***: How does reviewers' sentiment toward different aspects change across rounds of review?

This entails exploring the fluctuation in positive and negative sentiment distribution for various fine-grained aspects throughout the multi-turn review process.

***RQ3***: Is there a correlation between sentiment scores of review aspects and the final number of review rounds? Which review aspect shows stronger correlation?

This analysis aims to identify specific review aspects such as experiments, results analysis, etc. that reviewers focus on, potentially influencing the duration of review cycles. Understanding these correlations can offer writing guidance for authors.

The above work utilizes large-scale data for text analysis to explore the distribution and changes of sentiment in different rounds of review comments. Theoretically, this study can explore the potential characteristics and patterns of the process of multi-round peer reviews, which enrich the research content of review comment content mining and expand the research extension of sentiment analysis. In terms of application value, the research results of our research can help authors to understand the key aspects that affect the number of review rounds, such as experiments, results analysis, etc. It can also guide authors to avoid problems in these aspects when writing and revising papers, otherwise it is likely to lead to the extension of the review cycle.

[1] Faculty Opinions. Identifying highly active anti-CCR4-CAR T cells for the treatment of T-cell lymphoma [EB/OL]. [2023-05-04]. https://facultyopinions.com/article/742611967

All the data and source code of this paper are freely available at the GitHub website https://github.com/RuxueHan/Aspect-Sentiment-Analysis-of-Peer-Review-in-NC/blob/main/Annotation-Specification.md.

## 2 Related Work

In this section, we provide an overview of relevant research in two key areas: peer review comment text mining and fine-grained aspect sentiment analysis.

### 2.1 Overview of peer review comment mining research

Peer review reports contain evaluations, questions and suggestions provided by reviewers concerning the research content. Utilizing open-access peer review comments can reveal patterns and trends in content, structure, and sentiment of the reviews. Kang et al. (2018) proposed the first publicly available scientific peer review dataset PeerRead for research, revealing that certain paper characteristics, such as the presence of appendices, may be correlated with a higher acceptance rate. Wang & Wan (2018) were the first to investigate the task of automatically predicting overall recommendation decisions. Their findings indicated that, apart from outlier reviews, there was generally good consistency between the review texts and the final decisions. Gao et al. (2019) explored the rebuttal phase's impact on final scores, highlighting influences of initial scores and differences with other reviewers. Chakraborty et al. (2020) conducted an aspect-based sentiment analysis on the ICLR review dataset. They found that sentiments related to aspects such as robustness of the empirical/theoretical impact of ideas, and clarity were the most critical factors in predicting the final recommendation decisions in reviews. Zhang et al. (2022) found that reviewers with different genders, economic backgrounds, and cultural backgrounds tended to produce reviews of different lengths. Sun et al. (2023) explored how reviewers' academic status relates to the linguistic and sentiment features of their reviews. The research revealed that when reviewers were aware that their reputation exceeded that of the corresponding authors, there were no significant differences in the sentiment features of their comments. However, reviewers with higher reputations tended to use longer and more complex words in their reviews.

Relevant scholars are also paying attention to the field of automatic review and exploring it. Automated scholarly paper review (ASPR) refers to the process of computer or other intelligent machines independently reviewing the content of academic papers and automatically generating review reports. Lin et al. (2023) proposed the concept and process of Automated Scholarly Paper Review (ASPR). The various phases of ASPR, such as the annotated corpus set construction(Yuan et al., 2022), text parsing, originality measurement (Hou et al., 2020), novelty evaluation (Wang et al., 2024), and the generation of review reports, have been studied and preliminarily implemented. However, there are also some limitations, such as the bias caused by the data imbalance caused by the non-disclosure of the data of the rejected papers, and the defects of deep logical reasoning.

Additionally, some scholars have integrated peer review with scientometrics to explore the relationship between peer review and citations. Li et al. (2019) employed a neural prediction model based on peer review data to predict citation frequency, highlighting the role of peer review comments in this prediction. Ni et al. (2021), using Nature Communications as an example, employed a linear regression model and found no significant citation advantage for open peer review articles compared to non-open peer review articles. And they discovered no correlation between the number of review rounds and citation frequency. Qin et al. (2023) found that the distribution of peer review comments in different paper structures was not related to citation frequency.

**Table 1 Overview of research related to peer review comment mining**

| Authors | Scale of Corpus | Sources | Domains | Main Finding |
|---|---|---|---|---|
| Kang et al., 2018 | 10,770 Reviews | NIPS, ICLR, ACL,CoNLL | Deep Learning, Computational Linguistics, Natural Language Processing | Certain characteristics of a paper, such as the presence of an appendix, may be associated with a higher acceptance rate. |
| Wang & Wan, 2018 | 4,392 Reviews | ICLR | Deep Learning | Overall good consistency between review texts and final decisions, except for boundary reviews. |
| Gao et al., 2019 | 4,054 Reviews 1,227 Responses | ACL | Computational Linguistics | Reviewer's final score depends heavily on the difference between their initial score and the initial scores of other reviewers. |
| Li et al., 2019 | 2,123 Articles' abstract & 8,290 Reviews | ICLR, NIPS | Machine Learning, Computational Neuroscience | Peer review comments can play a role in citation prediction tasks for academic papers. |
| Chakraborty et al., 2020 | 8,151 Reviews | ICLR | Deep Learning | The distribution of aspect sentiment is significantly |

| | | | | different between accepted and rejected papers. |
|---|---|---|---|---|
| Ni et al., 2021 | 7,614 Articles & 2,293 Reviews | Nature Communications | Biological sciences, Health Sciences, Physical Sciences, Earth and Environmental Sciences, sociology | There were no significant citation advantage results for open peer-reviewed papers compared to non-open peer-reviewed papers. |
| Sun et al., 2023 | 2580 Reviews | eLife | Life and biomedical sciences | Reviewers had no apparent bias toward academic standing, while more prestigious reviewers tended to use longer, more complex words in their comments. |

Table 1 summarizes research on text mining of peer review comments. In conclusion, there is a growing interest among scholars in exploring text mining in the context of peer review comments. However, current research often lacks focus on the characteristics of multi-round review processes. Additionally, issues related to the broad granularity of aspects may limit the value provided for authors to optimize their submissions.

## 2.2 Overview of the aspect-based sentiment analysis research

Aspect-based sentiment analysis generally involves two main tasks: aspect extraction and aspect-level sentiment classification. Qiu et al. (2011) introduced the Double Propagation (DP) algorithm, which combined with syntactic dependency trees to simultaneously extract sentiment words and opinion targets. Dong et al. (2014) proposed an adaptive recurrent neural network model (AdaRNN) that utilizes inter-word dependency rules and recurrent neural networks to represent the vectors of target aspect words and their context in text. Wang et al. (2016) introduced the Aspect Embedding LSTM and Attention-based LSTM to make more effective use of aspect information. They then combined these two components to create the Attention-based LSTM with Aspect Embedding(ATAE-LSTM) model, which employs an attention mechanism for target embedding and is applied to aspect sentiment classification tasks. Recent state-of-the-art models, leveraging BERT's robust capabilities in transfer learning and contextual modeling. Song et al. (2019) built an Attention Encoder Network-based model called AEN-BERT. This network is designed to capture hidden states and semantic interactions between target aspect words and context words. Zeng et al. (2019) introduced the LCF-BERT model, emphasizing the impact of words closer to the aspect word on its final sentiment orientation. In addition to utilizing global contextual information, this model defines a local context processor specifically for extracting local features of aspect context. With the advancement of Explainable Artificial Intelligence (XAI), increasing attention has been directed toward enhancing the interpretability of classification models in natural language processing (NLP) tasks to foster better understanding and trust in AI systems. In this context, Dehdarirad (2025) proposes a unified framework for evaluating explainability methods in language classification models.

There have been some studies conducted by scholars on sentiment analysis of peer review comments. Wang & Wan (2018) focused on academic articles' sentiment analysis in peer review comments, introducing a Multiple Instance Learning Network with an Abstract-based Memory mechanism (MILAM). Chakraborty et al. (2020) constructed an aspect-based review corpus using an active learning framework. They found that a Universal Sentence Encoder embedding Feed-Forward Neural Network (FFNN-uni) model performed the best in aspect-level sentiment classification tasks. Their analysis accepted papers generally had significantly higher positive sentiment scores. Ghosal et al. (2019) developed a deep neural network that incorporates information from paper text, review text, and associated sentiment. This model predicts the acceptability and recommendation scores for research papers, demonstrating a significant improvement with the inclusion of sentiment from reviews. Thelwall et al. (2020) introduced PeerJudge, a sentiment analysis program determining reviewers' decisions based on praise and criticism in peer review reports. Kumar et al. (2022) proposed a novel deep neural architecture, showing improved performance of peer review decision prediction by incorporating aspect information and their corresponding sentiments. Furthermore, the study found that sentiments related to aspects such as motivation, meaningful comparisons, originality, and authenticity were the most effective in predicting recommendations.

**Table 2 Overview of research on sentiment analysis of peer review comments**

| Authors | Scale of Corpus | Sources | Domains | Main Finding |
|---|---|---|---|---|
| Wang & Wan, 2018 | 4,392 Reviews | ICLR | Deep Learning | Multiple Instance Learning Network with a novel Abstract-based Memory mechanism (MILAM) |
| Ghosal et al., 2019 | 10,325 Reviews | ICLR, ACL, CoNLL | Deep Learning, Computational Linguistics, Natural Language Processing | Deep neural architecture — DeepSentiPeer |
| Chakraborty et al., 2020 | 8,151 Reviews | ICLR | Deep Learning | Universal sentence encoder embedding FFNN(FFNN-uni) |
| Thelwall et al., 2020 | 7,333 Reviews | F1000, ICLR | Biomedicine | Dictionary-based sentiment analysis program—PeerJudge |

| Kumar et al., 2022 | 28,119 Reviews | ICLR, NeurIPS | Deep Learning, Artificial Intelligence | Deep neural architecture — DeepASPeer |
|---|---|---|---|---|

Table 2 displays related research on sentiment analysis of peer review comments. In conclusion, as peer review comment corpora become more available, there is a growing focus on analyzing the content of review texts and extracting sentiment information. However, there are still challenges to be addressed regarding features related to the review rounds, fine-grained aspect extraction, and fine-grained sentiment analysis in the context of peer review.

A closely related work is by Han et al. (2022), where they utilized the DP algorithm for aspect extraction from peer review comments with multiple rounds from Nature Communications. Han et al. further identified fine-grained review aspect categories and investigated how the attention to different aspects changed across rounds of review. Unlike that work, this paper takes a further step by exploring fine-grained aspect sentiment in multi-round reviews, analyzes the distribution and variations of aspect sentiments across rounds, and investigates the correlation between aspect sentiment in review comments and the number of review rounds. Another related work is Zhou et al. (2024), which primarily investigates the impact of aspect-level reviewer sentiments on the duration of the peer review process, with an emphasis on enhancing review efficiency. In contrast, the present study focuses on the temporal evolution of reviewer sentiments across multiple review rounds and explores their correlation with the number of review iterations. Methodologically, this work employs a broader range of sentiment classification models, utilizes a different dataset partitioning strategy, and achieves superior classification performance. While Zhou et al. (2024) emphasizes practical implications for review time prediction, this study centers on modeling dynamic sentiment trajectories.

## 3 Methodology

In this study, we use peer review comments for accepted papers in Nature Communications as data source. We aim to analyze the aspect sentiment distribution and variations across multiple review rounds to understand reviewers' focus. Additionally, we explore the relationship between review aspects’ sentiments and review rounds. The main research contents comprise three parts. First, the training data set is constructed by manually labeling the aspect sentiment polarity in the review comment. Then we conduct comparative experiments employing various aspect sentiment classification models by deep learning method, selecting the best-performing model for automatic sentiment classification to apply in the entire review comments corpus. Finally, a statistical analysis examines the distribution of aspect sentiment in multi-round review comments and explores how it changes with each round. Sentiment scores are calculated at both the aspect and document levels. And this study explores the correlation between sentiment scores for various review aspects and the number of review rounds. The study framework is shown in Figure 2.

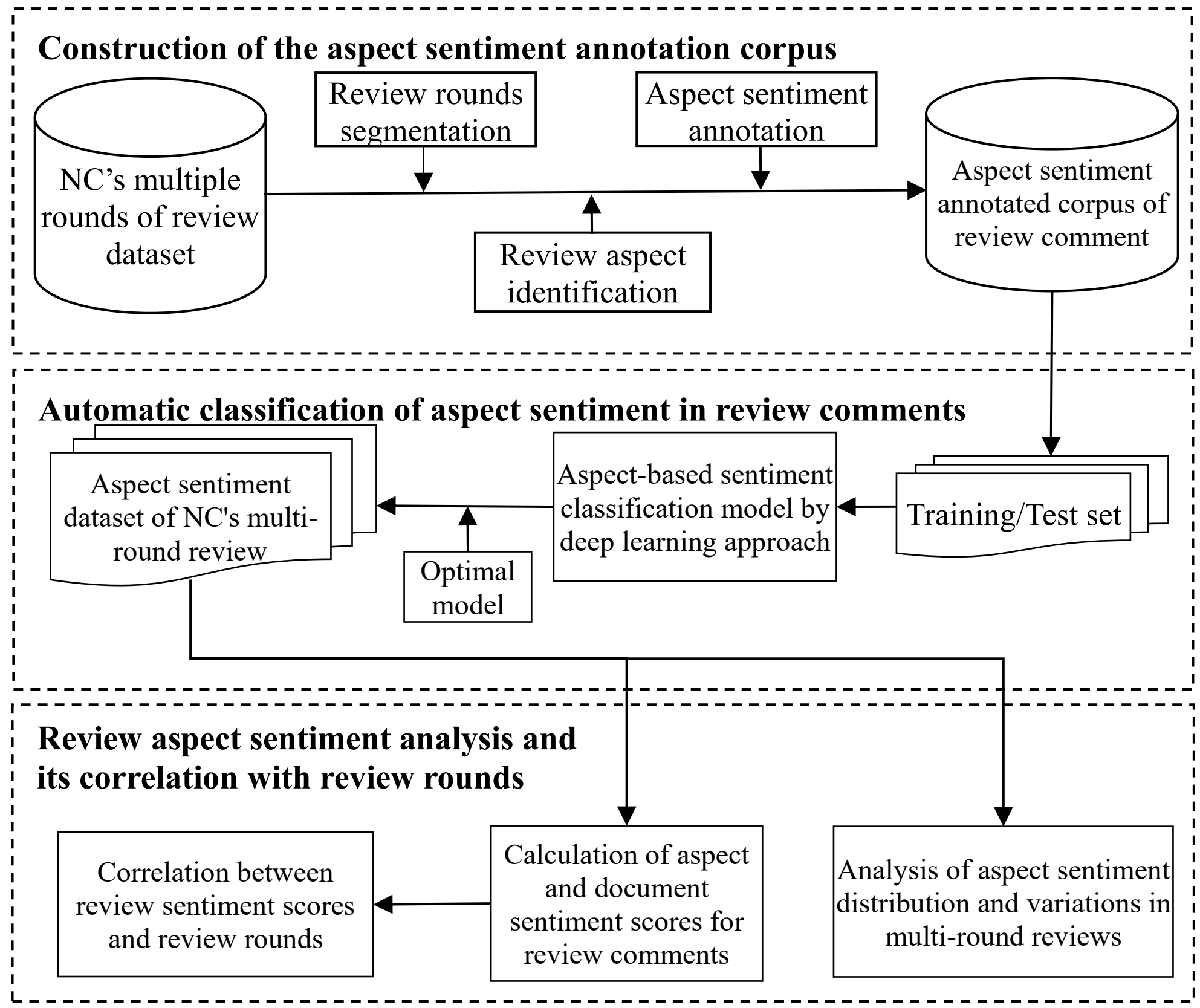

**Figure 2 The framework of this study**

## 3.1 Overview of multiple rounds of peer review comments datasets

### (1) Collection of review corpus

This paper collects peer review comments for papers published in Nature Communications (hereinafter referred to as NC) journals from 2016 to 2020. A total of 41,373 peer review files corresponding to 11,063 accepted papers are used as the experimental data for constructing the review comments corpus.

NC is a highly influential interdisciplinary, which has implemented a Transparent Peer Review format since January 2016. Under this format, the journal uploads records of multiple rounds of peer review comments and author response letters in PDF format alongside the published paper. Therefore, the NC journal provides a substantial amount of open peer review data covering various disciplines, meeting the research requirements for conducting aspect sentiment analysis of multi-round peer review comments in this paper. Table 3 presents the statistical results of the collected review corpus.

**Table 3 Statistical results of the peer review comments**

| First-level disciplines | Number of second-level disciplines included | Number of peer review files | Percentage of peer review files |
|---|---|---|---|
| Biological sciences | 24 | 21683 | 52.41% |

| Earth and environmental | 12 | 2588 | 6.25% |
|---|---|---|---|
| Health sciences | 17 | 6090 | 14.72% |
| Physical sciences | 9 | 10180 | 24.61% |
| Scientific community and society | 9 | 832 | 2.01% |
| Total | 71 | 41373 | 100% |

**(2) Text preprocessing of the review comments**

Firstly, this study segments and stores peer review documents based on the review rounds by employing structural rule recognition and manual annotation. Some review texts explicitly mark the start of a new round with phrases like "Reviewer Comment," enabling us to uniformly apply structural rule recognition for segmentation. For review texts without such markers, three students majoring in Information Management add labels (@@Round-i, i=1,2,...,6) to identify review rounds through manual judgment. Then, we conduct text preprocessing on all the review comments after dividing rounds, clean the noise values such as scrambled and illegal characters, unify the coding format, and use the send_tokenize[2] tool for sentence processing, and a structured corpus of review comments with multiple rounds is constructed to facilitate subsequent experimental studies by review sentences (Han et al., 2022).

Table 4 presents statistics on the number of review comments with different review rounds in each discipline. The results indicate that accepted papers in NC underwent between one and six review rounds. Among the five first-level disciplines, most papers underwent two rounds of review, followed by 3 rounds. Approximately 90% of the NC's accepted papers experienced 2-3 rounds of peer review from submission to recruitment.

**Table 4 Statistics on the number of review comments with different review rounds in each discipline**

| First-level disciplines | Number of review comments with different review rounds/pieces （proportion/%） | | | | | |
|---|---|---|---|---|---|---|
| | 1 Round | 2 Rounds | 3 Rounds | 4 Rounds | 5 Rounds | 6 Rounds |
| Biological sciences | 1721(8.10) | 15131(71.19) | 3976(18.71) | 376(1.77) | 46(0.22) | 3(0.01) |
| Earth and environmental sciences | 158(6.23) | 1625(64.13) | 630(24.86) | 100(3.95) | 20(0.79) | 1(0.04) |
| Health sciences | 418(6.97) | 4135(68.98) | 1283(21.40) | 145(2.42) | 12(0.20) | 2(0.03) |
| Physical sciences | 801(8.05) | 7153(71.88) | 1748(17.56) | 205(2.06) | 38(0.38) | 7(0.07) |
| Scientific community and society | 42(5.17) | 551(67.86) | 194(23.89) | 23(2.83) | 2(0.25) | 0(0) |
| Total | 3140(7.75) | 28595(70.53) | 7831(19.31) | 849(2.09) | 118(0.29) | 13(0.03) |

## 3.2 Aspect identification of the review comments

[2] https://www.nltk.org/api/nltk.tokenize

First, the spaCy[3] toolkit is utilized to identify noun phrases in the review text. These phrases are connected with '_' in their original sentences, forming new compound nouns.

Then, the DP (Double Propagation) algorithm proposed by Qiu et al. (2011) is employed to extract aspect words from peer review comments. Utilizing Stanford-CoreNLP for part-of-speech tagging and syntactic dependency parsing, aspects are considered nouns and opinions as adjectives. Dependency relations between opinion words and aspects are restricted to types such as mod and obj, while relations between opinion words or aspects themselves are limited to conj. Candidate aspect words are extracted using the DP algorithm.

Finally, the top 10% of candidate aspect words based on word frequency are composed a high-frequency aspect word set. The Affinity Propagation (AP) method (Frey & Dueck, 2007) is employed for clustering these high-frequency aspect words to group different aspect words expressing the same meaning. In this process, 200-dimensional word vectors are constructed for each review aspect word using the Continuous Bag of Words (CBOW) mode of Word2Vec (Mikolov et al., 2013). Cosine similarities between review aspect words are calculated and fed into the AP clustering model. The resulting clusters of review aspects are defined based on the meanings of the different clusters as determined by the experimental results, thereby determining the final clusters of review aspects (Han et al., 2022). The specific categories are presented in Table 5.

**Table 5 Aspect Clusters of Peer Review Comments**

| **Cluster** | **Aspects（Proportion/%）** |
|---|---|
| Problem Definition/Idea**(PDI)** | question(8.38)/issue(7.09)/ problem(4.69)/view(3.09)/ assumption(2.42)/definition(2.39)/ concept(2.92)/direction(1.50)/... |
| Data/Datasets**(DAT)** | data(57.26)/information(10.50)/ sample(4.81)/signal(4.28)/ dataset(2.15)/material(3.26)/... |
| Tables & Figures **(TNF)** | figure(51.08)/fig(12.37)/ table(7.46)/panel(3.34)/ legend(2.77)/image(2.44)/ plot(1.78)/label(1.10)/... |
| Methodology & Model**(MET)** | model(20.26)/structure(10.65)/ approach(8.63)/method(8.85)/ mechanism(7.63)/strategy(1.80)/... |
| Experiments & Operations**(EXP)** | control(10.67)/experiment(7.33)/ test(7.19)/process(4.40)/ testing(1.90)/processing(1.43)/... |
| Conclusions & Discussions**(CON)** | discussion(19.61)/claim(12.92)/ conclusion(8.08)/explanation(7.79)/ finding(6.80)/discovery(2.21)/... |

[3] https://spacy.io/

| | |
|---|---|
| Index/Parameters **(IND)** | state(8.44)/level(4.92)/type(4.31)/ size(4.19)/rate(3.34)/range(3.26)/ standard(2.60)/concentration(2.54)/ population(2.49)/threshold(1.06)/... |
| Result Analysis & Evaluation**(RES)** | analysis(23.11)/result(7.89)/ measure(3.35)/estimation(2.67)/ evaluation(1.33)/accuracy(1.18)/ calculation(1.03)/... |
| Structure & Language**(SNL)** | expression(18.02)/statement(5.54)/ understanding(5.38)/description(5.02)/ paragraph(4.31)/context(3.66)/ clarity(0.88)/... |
| Impact & Research Value**(IMP)** | effect(11.65)/mean(5.93)/ impact(5.15)/value(4.17)/ contribution(3.17)/performance(2.42)/ influence(2.39)/novelty(1.77)/... |
| Comparison & Correlation**(CMP)** | difference(10.67)/comparison(10.11)/ correlation(6.62)/association(4.65)/ contrast(2.67)/similarity(0.93)/... |

### 3.3 Aspect-based sentiment analysis by deep learning approach

This section introduces the research method of constructing the corpus of aspect sentiment annotation and the automatic classification of aspect sentiment.

#### 3.3.1 Construction of aspect sentiment corpus of review comment

First, a detailed set of guidelines is formulated for annotating aspect-level sentiment in review comments. These guidelines precisely define the criteria for judging neutrality, positivity, and negativity of aspects. Annotators undergo training to ensure consistent and rational annotation. The complete guidelines are available at accessed at https://github.com/RuxueHan/Aspect-Sentiment-Analysis-of-Peer-Review-in-NC/blob/main/Annotation-Specification.md.

In the formal annotation process, three annotators follow the annotation guidelines and judge the sentiment polarity of aspect words in review sentences and label them with the corresponding tags. Sentiment polarity labels included '1' (positive), '-1' (negative), '0' (neutral), and '2' (ambiguous). If the sentiment of an aspect word couldn't be determined or the word isn't an aspect in the given review sentence, it is labeled as ambiguous sentiment. An example of the annotation is shown in Table 6.

**Table 6 Sample of the sentiment polarity annotation of the aspects in review sentences**

| Opinion words | Syntactic relationship | Aspect words | Label polarity | Review sentence |
|---|---|---|---|---|
| important | amod | finding | 1 | that would be a very important finding！ |
| previous | amod | figures | -1 | the re-referencing of previous figures out of order is confusing. |
| detailed | amod | analysis | 0 | detailed analysis of inoa scs (figure 5) suggests different maturation states which are then compared to stages a,b and c from analysis in figure 4b. |

To ensure annotated corpus quality, an independent cross-annotation approach is employed. The entire corpus is partitioned into three equal parts, with each portion independently labeled for aspect sentiment by two annotators. The Kappa coefficient (Warrens, 2011) calculates consistency between annotators, resulting in consistency test results of 0.63, 0.66, and 0.63. For inconsistent annotations, the third annotator marks the final annotation. In total, 5,063 review sentences with annotated aspect sentiments are collected, including 1,498 positive, 1,652 neutral and 1,913 negative aspect sentiment sentences.

#### 3.3.2 Deep Learning Method for aspect-based sentiment classification in review comments

In this paper, based on the annotated aspect sentiment corpus from NC peer review comments, we construct and train deep learning-based aspect sentiment classification models. These models are categorized into two groups based on whether they utilized the pre-trained BERT language model for text representation. Below are the basic principles of each model:

**(1) Deep Learning model not based on BERT text representation**

This paper first utilizes the GloVe model to represent the text data of peer review comments, using it as input for the deep learning models.

**LSTM-based sentiment classification model.** Long Short Term Memory (LSTM) is a widely used recurrent neural network model in NLP tasks. Tang et al. (2016) introduced the Target-Dependent LSTM (TD-LSTM) model. The TC-LSTM model builds upon TD-LSTM and further strengthens the association between the target word and every token in the sentence. Wang et al. (2016) introduced the ATAE-LSTM model, which learns a corresponding aspect embedding for each aspect. .

**MEMNET-based sentiment classification model.** This model splits word embeddings into aspect word embeddings and context embeddings. Context embeddings are stacked together and considered as an external memory (m). The model consists of several computational layers that include attention layers and linear layers. (Tang, Qin, & Liu, 2016).

**Interactive attention network (IAN).** IAN is based on the idea that both the target aspect and context words deserve special treatment. It learns its own representations by interactively modeling the target aspect and context, using LSTM networks to obtain hidden states of both the target aspect and context words.(Ma et al., 2017).

**Attention-over-attention Neural Networks (AOA).** AOA establishes a joint approach to build models for both aspect words and sentences, capturing the interaction between aspect words and contextual sentences. This enables joint learning between aspect and sentence representations, with automatic attention to important components of sentences (Huang et al., 2018).

**(2) Deep Learning model based on BERT text representation**

To fully utilize the context around the target aspect and extract more semantic information, this paper employs BERT-based deep learning model for aspect sentiment classification in peer review comments.

**BERT-SPC Model.** BERT-SPC is a model that utilizes the BERT pre-trained model for classifying pairs of review sentences. It involves inputting sequences like "[CLS]+context+[SEP]+target aspect+[SEP]" into the base BERT model to accomplish the sentence pair classification task (Devlin et al., 2019).

**AEN-BERT Model.** AEN-BERT is an Attention Encoding Network based on target sentiment classification. It utilizes a pre-trained BERT model and an attention-based encoder to model the relationship between the context of the review sentence and the target review aspect word (Song et al., 2019).

**LCF-BERT Model.** Zeng et al. (2019) proposed the LCF-BERT model based on the idea that words closer to the aspect word hold more influence on sentiment. This model uses word counts to measure the relative semantic distance between context words and the target review aspect word. It utilizes Local Contextual Focus (LCF) and Semantic Relative Distance (SeRD) to identify context words within the local context for a specific review aspect word.

**LCFS-BERT Model.** Phan & Ogunbona (2020) introduced the LCFS-BERT model, which improves upon the LCF-BERT model by incorporating the shortest path between two words in the syntactic dependency tree as the Syntactic Relative Distance (SeRD). This enhancement allows the aspect sentiment classifier to understand syntactic relationships between review aspect words and context words.

**Aspect sentiment classification model based on FT-RoBERTa.** Dai et al. (2021) conducted a comparison on three representative tree-based aspect-level sentiment classification models, ASGCN, PWCN, and RGAT. The induced trees generated from fine-tuned RoBERTa models tended to establish connections between sentiment words and other words.

## 4 Experiments and Results Analysis

This section focuses on the experiments of automatic review aspect sentiments classification, analysis of aspect sentiments in multi-round reviews, and correlation analysis between review sentiment and the number of review rounds. Detailed experimental procedures and result analyses for each section are presented below.

### 4.1 Analysis of automatic classification results of review aspect sentiments

To construct and experimentally test various aspect sentiment classification models, First we partition and preprocess the annotated data. Then, the best model is selected by performance comparison.

**(1) Experimental setup and result evaluation method**

The labeled data is divided into training, validation, and test sets at a ratio of 6:2:2. To ensure a balanced distribution of the experimental dataset, the labeled data for each of three sentiment polarities is initially split separately and then merged. Finally, text preprocessing involves replacing aspect words with "$T$" to meet model input requirements.

In training the deep learning model, the commonly used preset initial value in text classification tasks is selected initially, and then the hyperparameter setting of model is obtained after multiple rounds of tuning, as shown in Table 7:

**Table 7 Setting of models' hyperparameter**

| Hyperparameter | Interpretation | Setting Value |
|---|---|---|
| max_seq_len | The maximum length of a sequence | 85 |
| learning_rate | The step size at each iteration | 2e-05 |
| dropout | dropout ratio | 0.1 |
| batch_size | Size of batches of training | 16 |
| num_epoch | Number of iterations | 10 |
| Optimizer | Optimization algorithm | Adam |
| hidden_dim | Hidden layer dimension | 300 |
| bert_dim | Embedded dimension | 768 |
| SRD | Semantic relative distance | 3 |

This paper classifies sentiment polarity into three categories: positive, neutral and negative, so the aspect sentiment classification is essentially a multiple classification problem, we evaluate the model's performance using Accuracy, Macro-P, Macro-R and Macro-$F_1$ as the chosen metrics for the experimental results.

**(2) Performance comparison of aspect-based sentiment classification models**

The sentiment classification models are trained on NC's review corpus, and the experimental results are shown in Table 8：

**Table 8 Experimental results of each deep learning model**

| | Models | Accuracy (%) | Macro-P (%) | Macro-R (%) | Macro-$F_1$ (%) |
|---|---|---|---|---|---|
| Without Bert | LSTM | 68.81 | 68.98 | 68.98 | 68.92 |
| | TD-LSTM | 69.6 | 70.13 | 69.6 | 69.72 |
| | TC-LSTM | 67.62 | 67.71 | 67.88 | 67.66 |
| | ATAE-LSTM | 69.79 | 71 | 69.48 | 69.76 |
| | MEMNET | 71.37 | 72.8 | 70.89 | 71.15 |
| | IAN | 69.6 | 70.08 | 69.53 | 69.47 |
| | AOA | 68.61 | 68.6 | 69.04 | 68.77 |
| With Bert | AEN-BERT | 77.69 | 79.17 | 77.3 | 77.64 |
| | BERT-SPC | 81.05 | 81.99 | 80.76 | 80.85 |
| | **LCF-BERT-CDM** | **82.72** | **83.43** | **82.45** | **82.65** |
| | LCF-BERT-CDW | 81.15 | 82.32 | 80.77 | 81.02 |
| | LCFS-BERT-CDM | 82.33 | 82.75 | 82.17 | 82.31 |
| | LCFS-BERT-CDW | 81.54 | 81.78 | 81.5 | 81.53 |
| | ASGCN-RoBERTa | 77.99 | 78.32 | 77.96 | 78.01 |
| | PWCN-RoBERTa | 76.9 | 77.37 | 76.94 | 77.05 |
| | RGAT-RoBERTa | 78.87 | 80.38 | 78.35 | 78.74 |

In the aspect sentiment polarity classification task of peer review, BERT-based deep learning models outperform non-BERT models significantly. Experimental results indicate that BERT models effectively leverage the context of each word in the peer review text during the encoding process, thereby extracting richer semantic information. Among them, the LCF-BERT model, incorporating the CDM method, stands out with an Accuracy of 82.72% and a Macro-F1 score of 82.65%. Its success may be attributed to introducing semantic relative distance, emphasizing the impact of words closer to the aspect word on sentiment. The LCF-BERT-CDM model, which pays more attention to the local context words of specific aspects, is well-suited for the task of aspect sentiment polarity classification in peer review comments.

However, implicit sentiment expressions in peer review comment texts may lead to classification errors. Additionally, without considering the context of surrounding sentences, sentiment analysis may not accurately distinguish whether the reviewer is making factual statements about expert knowledge or evaluating the paper's content. These are potential reasons for sentiment polarity classification errors.

### 4.2 Aspect-based sentiment analysis of multiple rounds of review comments

In this study, the LCF-BERT-CDM model, which exhibits optimal performance, is employed for the automated classification of unannotated aspect sentiment corpora. The subsequent analysis focuses on the distribution of aspect sentiment and its changes with the review rounds of NC's review comments.

#### 4.2.1 Distribution of aspect sentiment in multiple rounds of review

We answer *RQ1* in this section. Heat maps are used to visualize the three sentiment polarity distributions of the top 10 frequency aspects in all the evaluation comments. The heatmap reveals that, in comparison to positive sentiments, there are significantly more instances of negative and neutral sentiments for these popular aspect words. This is because peer review often involves critiquing and making statements about the academic quality of research papers, leading to questions or suggestions about the core elements such as data, methods, and experiments in academic papers. The sentiment polarity frequency statistics reveal that aspect words like "data," "figure," "result," and "analysis" have the highest sentiment involvement. This indicates that reviewers pay particular attention to issues related to data, result analysis, and similar aspects during the review process, as data quality is crucial for the quality of research results, and figures related to data should accurately represent the information.

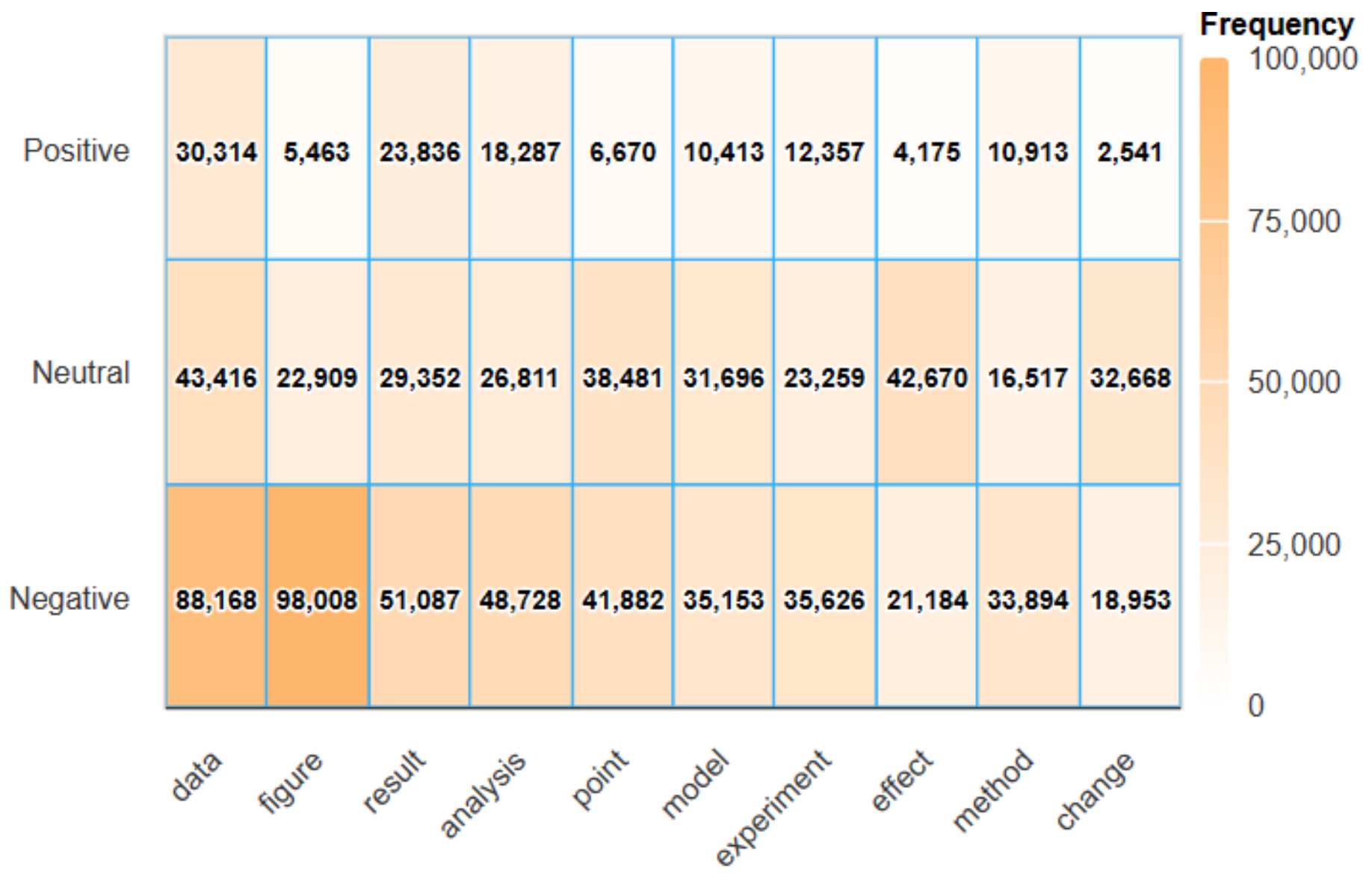

**Figure 3 Heat map of sentiment polarity frequency distribution of top10 aspect words**

Based on the established aspect classes, as detailed in Figure 4, the sentiment polarities of various aspect words within the same class are cumulatively combined. This study conducts a statistical analysis of the three sentiment polarities for various aspect classes under different review rounds. An example of a three-round peer review is presented in Figure 4(a)~(c):

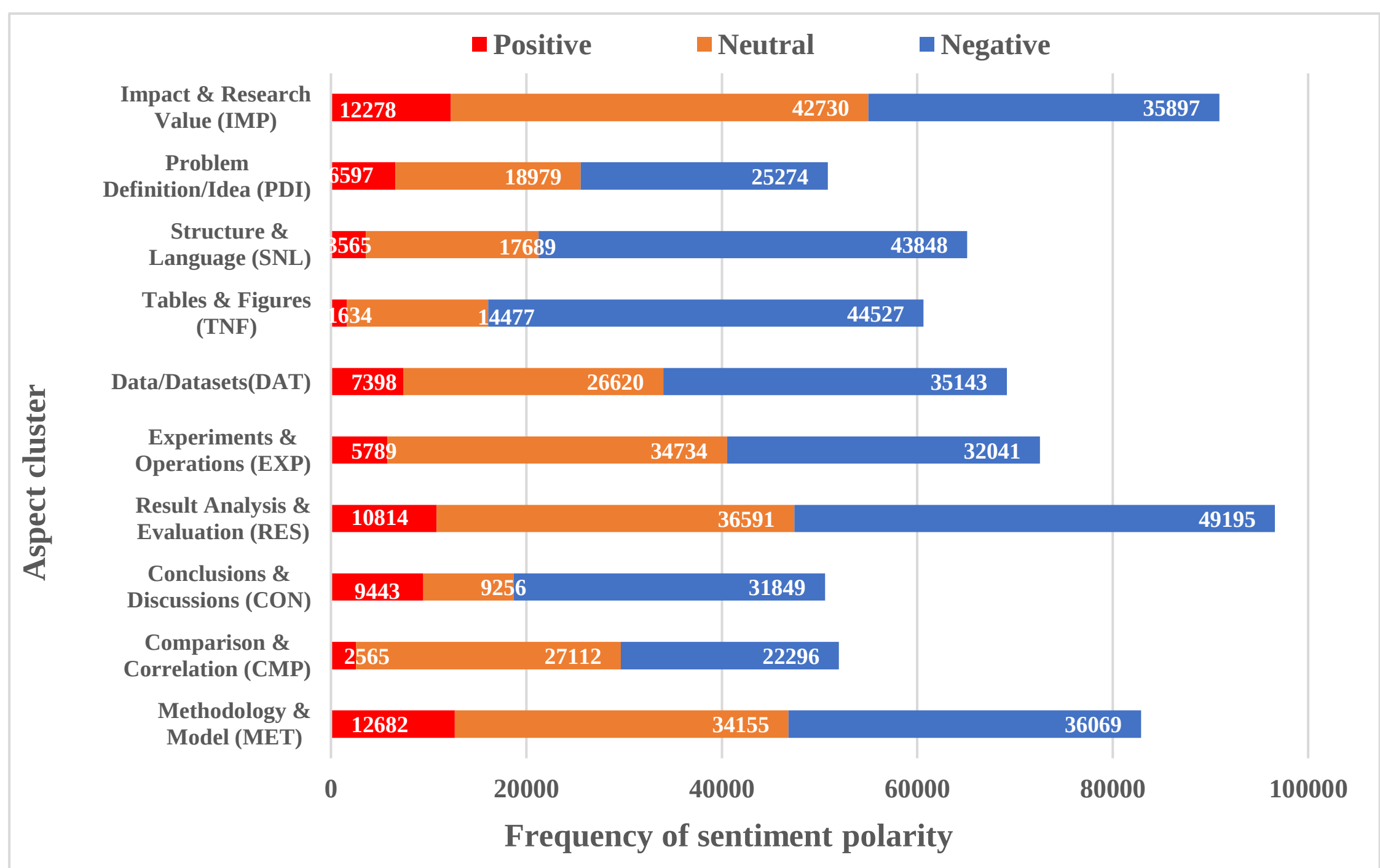

(a) Round=1

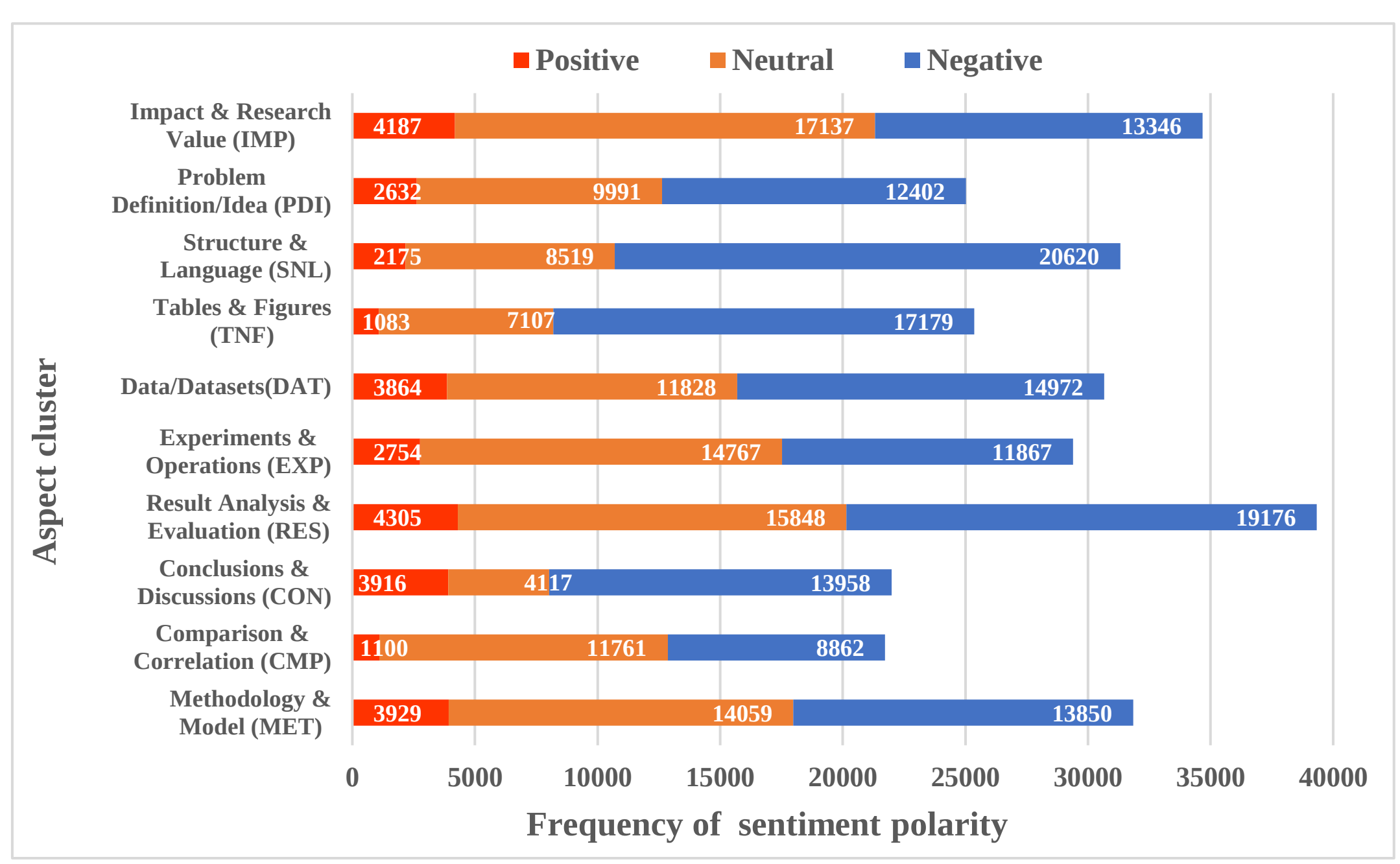

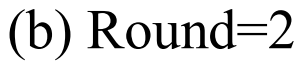

(b) Round=2

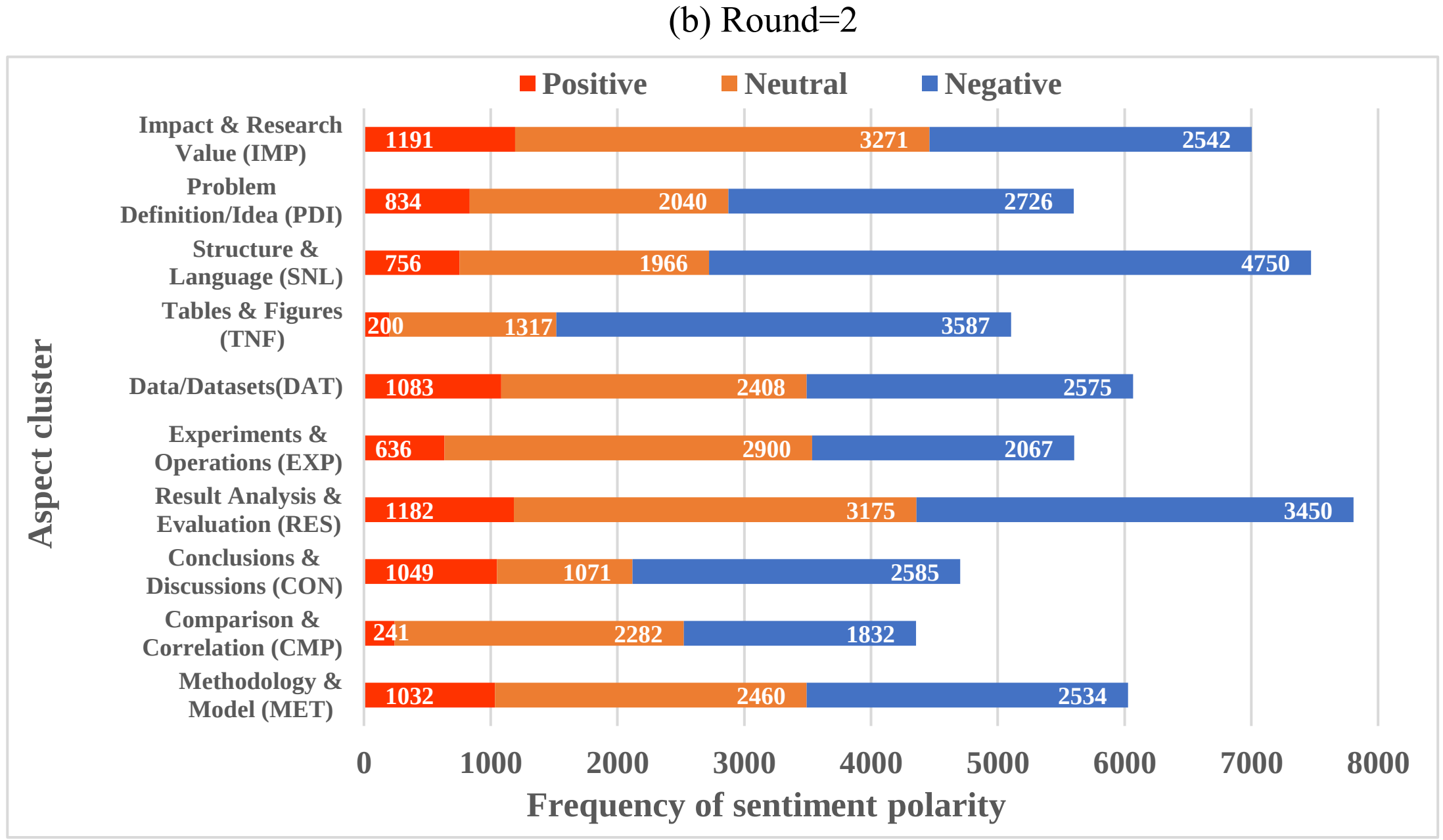

(c) Round=3

**Figure 4 The sentiment distribution of each aspect class under different rounds**

The results indicate that negative and neutral sentiments are predominant in the peer review comments in each round. As the number of review rounds increases, there is a diminishing distribution of sentiments associated with various aspect classes. This decline can be attributed to the reduction in comment length as the number of rounds progresses. Notably, aspect classes such as "Result Analysis & Evaluation(RES)," "Impact & Research Value(IMP)," and "Methodology & Model(MET)" consistently contain more sentiment content across all rounds. This suggests the enduring focus of reviewers on result analysis, research impact, value, and the reliability of methodologies and models throughout the review process. The result analysis section, serving as the foundation of the entire paper, emphasizes the fundamental and original purpose of presenting research findings. Research value and impact are the starting points and foundations of research work. The choice of appropriate methodologies and models is also at the core of research content, determining the effectiveness of the study. Consequently, this provides guidance to authors, suggesting that they should pay more attention to these aspects during the writing and revision process and not evade these issues when addressing reviewer comments. Neglecting them may lead to an extended review period. Additionally, there is an uptick in sentiment distribution of "Structure & Language(SNL)" aspect class in the final review round, indicating growing attention to language and other detailed aspects by reviewers as the review progresses. Reviewers in the early rounds tend to focus less on language and structural aspects of the paper. However, as key content issues are addressed and the review process is nearing completion, reviewers pay more attention to and evaluate the language and other finer details in the paper.

#### 4.2.2 The change of multiple rounds of review aspect sentiment

We answer *RQ2* in this section. Statistical analysis indicates that the majority of NC's accepted papers undergo either 2 or 3 review rounds. There is limited data available for peer review comments with more than 4 rounds, and single round reviews do not provide sufficient data for the analysis of multi-round aspect sentiment changes. Therefore, this paper focuses on analyzing aspect sentiment changes in reviews with a total of two or three rounds.

This paper explores the changing patterns of aspect sentiments in different rounds by calculating the percentage of positive and negative sentiment polarities in the total sentiments of each aspect class. The calculation formula is:

$$Aspect_{pos-i} = \frac{N_{pos-i}}{N_{pos-i} + N_{neu-i} + N_{neg-i}} \quad (1)$$

$$Aspect_{neg-i} = \frac{N_{neg-i}}{N_{pos-i} + N_{neu-i} + N_{neg-i}} \quad (2)$$

Where, $Aspect_{pos-i}$ and $Aspect_{neg-i}$ represent the positive and negative sentiment proportion of the aspect class "Aspect" in the i-th round of review comments respectively, $N_{pos-i}$, $N_{neu-i}$ and $N_{neg-i}$ respectively represent the frequency of positive, neutral and negative sentiments for the aspect class in the i-th round of review comments.

As shown in Figure 5(a)~(b), when the total rounds of review are two, with increasing rounds, there is a consistent trend across all aspect clusters: an increasing percentage of positive sentiments and a decreasing percentage of negative sentiments. This suggests that as authors optimize their papers based on review feedback, reviewers' sentiment tends to shift towards more positive. As depicted in Figure 5(c)~(d), when the total rounds of review are three, overall, with increasing rounds, most aspect clusters follow a similar trend, with an increasing percentage of positive sentiments and a decreasing percentage of negative sentiments. However, it's noteworthy that aspect classes like Methodology & Model(MET) and Conclusions & Discussions (CON) initially show a decrease in the percentage of positive sentiments followed by a significant increase from the second round to the third round. In contrast, the Tables & Figures (TNF) aspect class displays the opposite trend, with a rise in positive sentiment percentage followed by a decrease. This may indicate that during the first two rounds of review, reviewers are more critical and provide suggestions regarding models, conclusions, and other aspects related to the core content of the paper. In the final third round of review, there may be a greater focus on modifications and suggestions related to details such as tables and figures.

Regarding the magnitude of change, the Data/Datasets(DAT) class exhibits a relatively larger increase in positive sentiment distribution. This reflects that issues related to the paper's data are of heightened concern to reviewers and often have a significant impact on the total rounds of review. However, overall magnitude of change in the distribution of positive and negative sentiment percentages for each class is relatively small. This suggests that the reviewers' sentiment towards various aspects in different rounds of review does not significantly change compared to their sentiment in the first round of review. Therefore, authors need to ensure the quality of the initial version of their paper because the reviewers' sentiment attitude during the first round of review is likely to set the overall tone for multiple rounds of review.

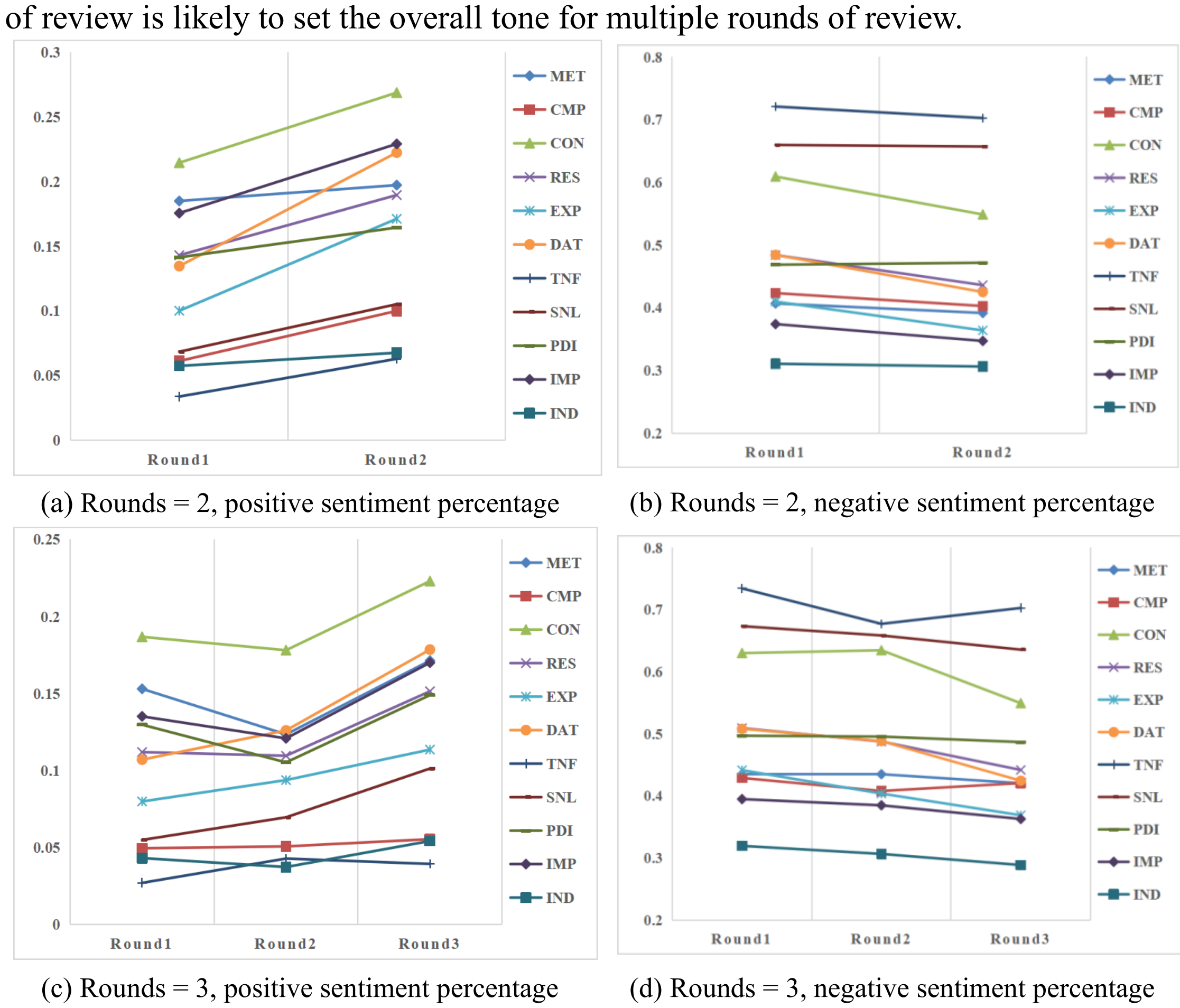

(a) Rounds = 2, positive sentiment percentage

(b) Rounds = 2, negative sentiment percentage

(c) Rounds = 3, positive sentiment percentage

(d) Rounds = 3, negative sentiment percentage

**Figure 5 The proportion of positive and negative aspect sentiment changes with the rounds**

(Han et al., 2023)

### 4.3 The correlation between review sentiments and number of review rounds

We answer *RQ3* in this section. In order to investigate which aspects that reviewers' sentiment towards in review comments affect the total number of review rounds a paper undergoes, this study conducted a correlation analysis between review sentiment scores and the total number of review rounds.

#### 4.3.1 Calculation of aspect-level and document-level sentiment scores in review comments.

This article calculates sentiment scores at two granular levels: aspect-level and document-level. The calculation rules for aspect-level sentiment scores are defined as follows: for each paper, the sentiment score for each aspect class is calculated as the difference between the frequency of positive and negative sentiment expressions in the corresponding peer review comments for that aspect class. This score is then standardized using the sum of the frequencies of all sentiment polarities for that aspect class in the review comments for that specific paper. The specific formula is as follows, where $Aspect_{score}$ represents the sentiment score for the aspect class "Aspect," and $N_{pos}$, $N_{neg}$, $N_{neu}$ are the frequencies of positive, negative, and neutral sentiment expressions for that aspect class in the review comments.

$$Aspect_{score} = \frac{N_{pos} - N_{neg}}{N_{pos} + N_{neu} + N_{neg}} \quad (3)$$

The document-level sentiment score for a peer review comment is calculated at the document level, considering the overall sentiment of the entire review document. In this case, the sentiment frequency for the entire document is obtained by summing up the sentiment frequencies for all aspects mentioned in that specific review. The calculation method aligns with the aspect-level sentiment score approach and can be expressed as follows:

$$Paper_{score} = \frac{M_{pos} - M_{neg}}{M_{pos} + M_{neu} + M_{neg}} \quad (4)$$

Among them, $Paper_{score}$ is the sentiment score of the Paper, $M_{pos}$, $M_{neg}$, $M_{neu}$ are the sum of positive, negative and neutral sentiment frequency of all aspects in the review comments.The calculation results for aspect-level and document-level sentiment scores for each paper are presented in Table 9:

**Table 9 Examples of sentiment score calculation results at document-level and aspect-level**

| Paper_ID | Review rounds | Document-level scores | Aspect-PDI score | Aspect-DAT score | Aspect-TNF score | Aspect-MET score | ... |
|---|---|---|---|---|---|---|---|
| s41467-020-14698-y | 2 | 0.1829 | 0.6667 | 0.2381 | 0.2857 | 0.1429 | ... |
| s41467-018-08153-2 | 3 | -0.5974 | -0.6667 | -0.6500 | -0.9697 | 0.1111 | ... |
| s41467-017-02268-8 | 4 | -0.1223 | -0.4444 | -0.3636 | -0.3636 | 0.1053 | ... |

This paper visualizes the document-level sentiment scores for all paper reviews using scatter plots, as shown in Figure 6. It indicates that as the total number of review rounds increases, the corresponding document-level sentiment scores become more concentrated toward the negative end. This suggests that papers that undergo more rounds of review before being accepted tend to receive a more negative overall sentiment from reviewers throughout the review process.

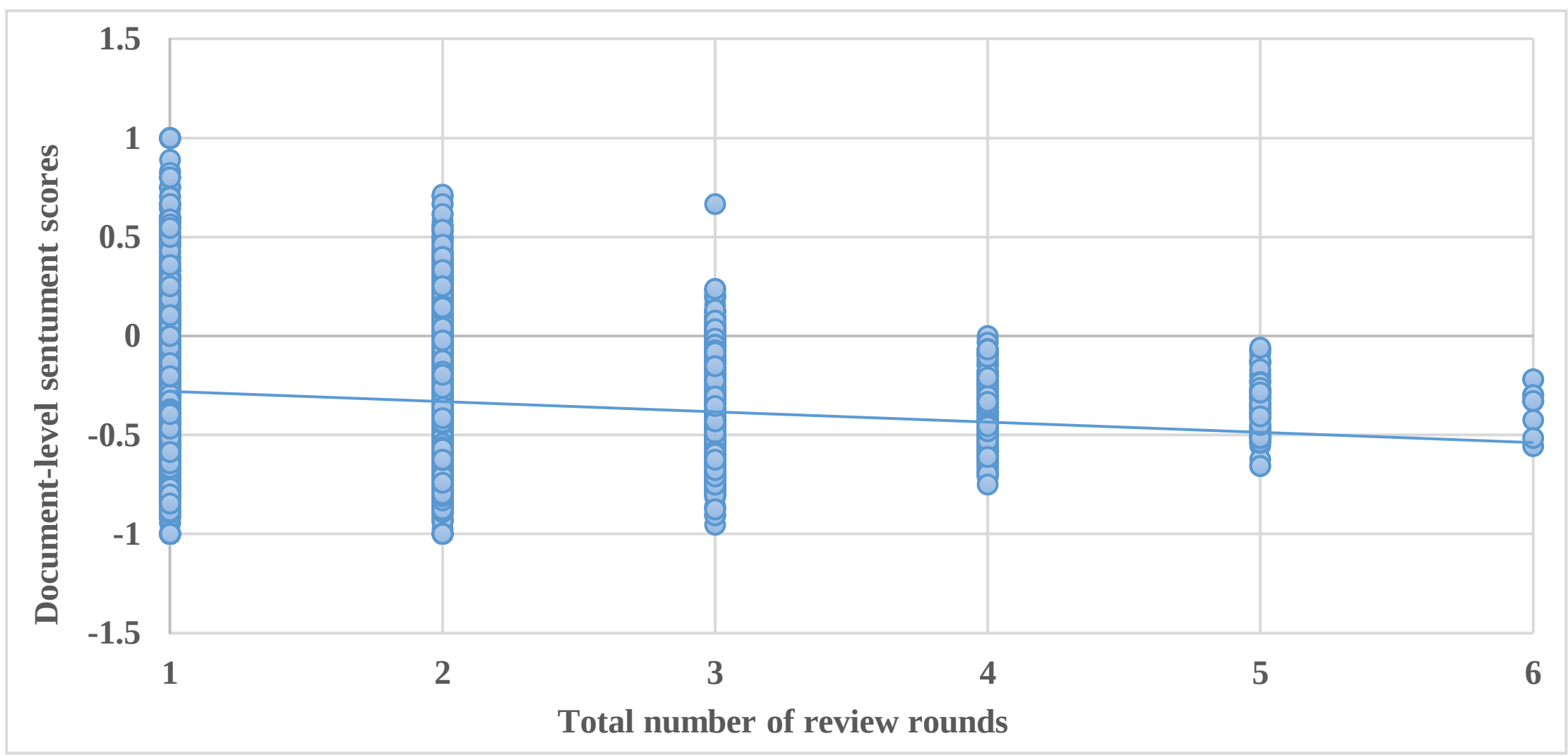

**Figure 6 Scatter plot of document-level sentiment scores**

### 4.3.2 Correlation analysis between review sentiment scores and review rounds

In this section, the results of the correlation analysis of the overall and each round's aspect sentiment scores with the number of review rounds are analyzed.

**(1) Correlation between the aspect sentiment score of overall multiple rounds and the number of review rounds**

First, a correlation analysis is conducted between the document-level sentiment scores and the number of review rounds. Results indicate a significant negative correlation and the Spearman correlation coefficient is -0.147, suggesting that papers receiving lower overall sentiment scores are more prone to undergo multiple review rounds. This result implies that when reviewers express a negative sentiment toward a paper, it is more likely to go through multiple rounds of review and revisions.

Next, from the aspect-level granularity, the results of the correlation analysis between sentiment scores of various aspect classes in the overall peer review comments and the number of review rounds are shown in Table 10. The table reveals a significant negative correlation ($p < 0.01$) between sentiment scores of all aspect classes and the number of review rounds. In other words, the lower the sentiment scores for aspect classes in the peer review comments of a paper, the more review rounds it is likely to undergo. Among these aspect classes, Experiments & Operations(EXP) and Impact & Research Value(IMP) exhibit the highest correlation with the number of review rounds, with correlation coefficients exceeding 0.15 in absolute value, suggesting a more substantial impact on the review process. On the other hand, Structure & Language(SNL) aspect classes have lower correlation coefficients and a relatively smaller impact on the number of review rounds. It's important to note that the data used in this study are peer review comments from accepted papers in the NC journal, introducing a potential survivorship bias. For example, the results of the correlation analysis may show a weaker correlation between the Methodology & Model(MET) aspect class and the number of review rounds, but this doesn't necessarily imply that this aspect has a minimal impact on the review rounds. It's possible that papers with very low sentiment scores in the Methodology & Model(MET) aspect have already been rejected and thus are not included in the study data. In summary, results indicate that negative evaluations of aspects such as experiments, results analysis, research significance, and data are more likely to result in additional review rounds and a prolonged review period. This insight also helps guide authors in prioritizing these critical aspects during paper writing, revision, and response to reviewer comments, aiming to minimize issues related to these aspects and thereby reduce the number of review rounds and shorten the paper's review period.

**Table 10 Correlation of the aspect sentiment scores of overall multiple rounds and the total number of review rounds**

| Aspect | Spearman correlation coefficient |
| --- | --- |
| Experiments & Operations, EXP | **-0.168**** |
| Impact & Research Value, IMP | **-0.163**** |
| Result Analysis & Evaluation, RES | **-0.157**** |
| Data/Datasets, DAT | **-0.149**** |
| Comparison & Correlation, CMP | -0.143** |
| Index/Parameters, IND | -0.139** |
| Conclusions & Discussions, CON | -0.138** |
| Problem Definition/Idea, PDI | -0.126** |
| Tables & Figures, TNF | -0.095** |
| Methodology & Model, MET | -0.098** |
| Structure & Language, SNL | -0.079** |

**Note:** ** is presented at 0.01 level (double tail) with significant correlation.

### (2) Correlation between the aspect sentiment scores of each round and the number of review rounds

Next, this paper further conducts a correlation analysis between the aspect sentiment scores of each round's review and the final number of review rounds. The results in Table 11 indicate that the correlation between the sentiment scores of review aspects in the first round and the number of review rounds is consistent with the correlation between overall sentiment scores of multi-round reviews and the number of review rounds discussed earlier. Specifically, aspects such as Experiments & Operations(EXP), Impact & Research Value(IMP) and Result Analysis & Evaluation(RES) exhibit a higher correlation with the number of review rounds. In the second and third rounds of review data, the correlation coefficients between the sentiment scores of various aspects and the number of review rounds are higher than in the first round. This suggests that after the first round of review and revision, if the reviewers' sentiment towards various aspects remains negative, the paper is more likely to undergo more review rounds. In these subsequent rounds of review, aspects such as Tables & Figures(TNF),

Conclusions & Discussions(CON) exhibit a higher correlation with the final number of review rounds, indicating that these aspects of the paper become more related to the total review rounds as the review process progresses. In the fourth round of review data, it's worth noting that the correlation between the sentiment scores of the Methodology & Model(MET) aspect and the number of review rounds is significantly higher compared to the previous rounds. This reflects that papers with very low sentiment scores in the Methodology & Model(MET) aspect may have been rejected in the early rounds of review. In the dataset of accepted papers, if reviewers have a more negative sentiment towards the Methodology & Model(MET) aspect, the paper is likely to undergo a longer review process. This indicates the significance of Methodology & Model(MET) as a crucial aspect influencing the number of review rounds.

**Table 11 Correlation of the aspect sentiment scores for each round with the number of review rounds**

| Aspect | Round1 | Round2 | Round3 | Round4 |
|---|---|---|---|---|
| Experiments & Operations, EXP | **-0.149**** | **-0.247**** | **-0.225**** | -0.163** |
| Impact & Research Value, IMP | **-0.143**** | -0.241** | -0.165** | 0.090 |
| Result Analysis & Evaluation, RES | **-0.143**** | **-0.277**** | **-0.250**** | **-0.246**** |
| Data/Datasets, DAT | **-0.135**** | -0.242** | -0.184** | -0.131* |
| Comparison & Correlation, CMP | -0.121** | -0.243** | -0.211** | 0.087 |
| Index/Parameters, IND | -0.124** | **-0.260**** | **-0.248**** | **-0.207**** |
| Conclusions & Discussions, CON | -0.108** | **-0.253**** | **-0.243**** | **-0.199**** |
| Problem Definition/Idea, PDI | -0.109** | -0.212** | -0.222** | -0.131* |
| Tables & Figures, TNF | -0.096** | **-0.251**** | **-0.229**** | -0.053 |
| Methodology & Model, MET | -0.071** | -0.239** | -0.213** | **-0.255**** |
| Structure & Language, SNL | -0.073** | -0.219** | -0.200** | 0.131* |

# 5 Discussion

## 5.1 Implication

Based on the constructed corpus of multiple rounds of reviews, this study uses natural language processing and machine learning techniques to conduct sentiment analysis on the fine-grained aspects of multiple rounds of reviews. The following will discuss the implications of this study from both theoretical and applied aspects.

**(1) Theoretical implication**

This study reveals two theoretical implications of research, including the necessity of focusing on the round characteristics of review comments with multiple rounds, and the advantages of fine-grained aspect sentiment analysis of review comments.

**Focus on the round characteristics of review comments with multiple rounds.** Currently, most research on mining review comments treats multi-round reviews as a whole and neglects the round information in peer reviews. The biggest difference between the study and previous studies is the mining of peer review comments in multiple rounds. In this paper, the review comments of journal articles are divided according to the review rounds, revealing the distribution and changes in aspect sentiments throughout the multi-round peer review process that academic articles go through from submission to publication. It found that there are differences in the focus and expressed sentiment tendencies of reviewers in each round of review. Therefore, it is necessary to pay attention to the round characteristics of peer review in the study of content mining of review comments.

**Advantages of fine-grained aspect sentiment analysis of review comments.** Currently, research on sentiment analysis of peer reviews often chooses aspects directly based on the standard guidelines of ACL conference reviews. However, these aspects, such as novelty and clarity, are typically broad assessments of the overall quality of the reviewed articles, which lack sufficient details to provide practical improvement suggestions for the authors. This study determines the fine-grained review aspect classes and conducts sentiment analysis of peer review comments at a more specific aspect level. The research results further confirmed that the review corpus contains rich sentiment information. Therefore, the application of the fine-grained sentiment analysis method to the content mining of peer review comments is helpful to explore the sentiment attitude of reviewers on the specific evaluation of the details of articles.

**(2) Practical implication**

The practical implications of this study primarily offer guidance to authors for improving their submissions. Exploring the the distribution and changes in aspect sentiments throughout the multi-round peer review process can help authors identify some aspects (e.g., results analysis, research value, methodology and model) that receive more attention from reviewer, and the impact of reviewers' emotional tendencies for each aspect on the total number of review rounds. The correlation analysis results show that the sentiments expressed by reviewers towards aspects such as Experiments & Operations(EXP), Impact & Research Value(IMP) are more strongly correlated with the number of review rounds. This can effectively help authors understand the important aspects that reviewers focus on during the review process for submitting papers to NC journals, which might affect the review cycle. It also provides valuable insights for authors to optimize and improve their articles, thus reducing the likelihood of extended review cycles.

## 5.2 Limitation

This study still has some limitations that need further improvement, including:

**(1) Issue of incomplete data coverage in experiments**

The experimental data in this study is limited to peer review comments for accepted papers, and comments for rejected papers are not included. There may be valuable aspect sentiment information in the comments for rejected papers, which could provide guidance to authors and help them avoid potential issues to improve the possibility of their paper acceptance. In addition, this article only selected NC journals as the research object to carry out experiments, not covering more journals and fields, and the research conclusions lack greater generality.

**(2) Performance of aspect sentiment classification model**

The performance of the aspect-level sentiment classification model needs improvement. Additionally, addressing the issue of implicit sentiment expression in review comments can be enhanced by incorporating domain knowledge and adding implicit sentiment lexicons.

**(3) Experimental errors of review aspect sentiment analysis**

To complete the aspect sentiment analysis across multiple rounds of review, this study involved various experimental processes. Each of these steps using machine methods may introduce some degree of error, and the final analytical results could be influenced by the accumulation of such errors, there also may be errors of human judgment in the annotation process. In this study, the results have been controlled by manual inspection and correction after each session as much as possible, but the final analytical results may still be affected by the accumulation of errors. Additionally, the experiments investigating the correlation between review sentiment and the number of review rounds do not incorporate control variables. In the future, we intend to further quantitatively assess and correct the experimental results through comparative experiments in the future so as to ensure the reliability of the findings.

## 6 Conclusion and Future Work

In this study, peer review comments from articles published in Nature Communications between 2016 and 2020 are served as the research dataset. Firstly, the corpus of aspect sentiment annotation of review comments with multiple rounds is constructed. The LCF-BERT-CDM model performs the best in aspect sentiment classification, achieving a Macro-$F_1$ score of 82.65%. The subsequent statistical analysis reveals that negative and neutral sentiments predominated in the peer review comments, while the proportion of positive sentiments in each aspect class increased as the number of review rounds increased, and the proportion of negative sentiments decreased. Furthermore, through a correlation analysis, it is found that the sentiment scores of various aspect clusters are negatively correlated with the total number of review rounds. Specifically, aspects related to experiments, results analysis, and research significance have a higher correlation with the total number of review rounds. For papers that underwent more rounds of review (four or more rounds), the sentiment scores of the methodology/model aspect class in the later review rounds are more strongly correlated with the total number of review rounds. This insight can guide authors to pay more attention to these review aspects when writing their papers to avoid prolonging the review process.

Future research should address current limitations by: expanding the size of the corpus to cover more journals and fields, so as to more comprehensively analyze the review process; using some State-of-the-arts models to improve the accuracy and reliability of aspect extraction and aspect-level sentiment classification. We will further explore the implied information from the reviewer differences and discipline differences, and the research work can be further expanded in the future.

## Acknowledgment

This study is supported by the Open Funding Project of Laboratory of ISTIC-Springer Nature Joint Laboratory for Open Science (Grant No. ISN23012).